# Tangles

*A structural approach to artificial intelligence in the empirical sciences*

---

Reinhard Diestel

**This ArXiv post**. . .

. . . contains the introductory Part I of the book with the above title. The print edition will appear in 2024 with Cambridge University Press.

An enhanced eBook edition, and open-source software, are available now at

`tangles-book.com`.

*Publisher's blurb*:

Tangles offer a precise way to identify structure in imprecise data. By grouping qualities that often occur together, they not only reveal clusters of things but also types of their qualities: types of political views, of texts, of health conditions, or of proteins. Tangles offer a new, structural, approach to artificial intelligence that can help us understand, classify, and predict complex phenomena.

This has become possible by the recent axiomatization of the mathematical theory of tangles, which has made it applicable far beyond its origin in graph theory: from clustering in data science and machine learning to predicting customer behaviour in economics; from DNA sequencing and drug development to text and image analysis.

Such applications are explored here for the first time. Assuming only basic undergraduate mathematics, the theory of tangles and its potential implications are made accessible to scientists, computer scientists and social scientists.

## Part II. Tangles in Different Contexts
*A collection of informal examples*

## Part III. The Mathematics of Tangles
*Concepts, theorems, algorithms*

# Preface

This is a book about potential applications of a new mathematical theory, written by a mathematician for a non-mathematical readership. Its style develops from an intuitively informal to a more formal level that uses basic mathematical language, just enough to make things precise. No serious mathematics is used anywhere in the main body of this book.[1]

This preface says a little about where tangles come from in mathematics, so as to indicate what is new in this book and what is not. Readers without this background are encouraged either to just skim the preface for a quick impression, or to skip straight ahead to Chapter 1. This begins with three separate introductions addressing natural scientists, social scientists, and computer scientists in turn.

The mathematical theory of tangles has its origins in the theory of graph minors developed by Neil Robertson and Paul Seymour in the final two decades of the 20th century. In a series of over twenty research papers, which culminated in the proof of one of the deepest theorems in graph theory, the *graph minor theorem*, Robertson and Seymour developed a new connectivity theory tailored specifically to the somewhat 'fuzzy' notion of their central object of study, that of a graph minor. Their new connectivity theory centred around a revolutionary new concept of high local connectivity in a graph: the notion of a *tangle*.

Loosely speaking, a tangle is a region of a graph that hangs together in an intricate way. Intricate in that, while being close-knit in the sense of being difficult to separate, it does not conform to the usual graph-theoretic notions of high connectivity.

Tangles constituted a shift of paradigm in what high *local* connectivity, somewhere in a graph or network, might mean. There is a standard measure of global connectivity for graphs, and the traditional way to measure their local connectivity was simply to look for regions in the graph that were highly connected in this global sense, applied to the region as a subgraph. As these highly connected regions were themselves viewed as graphs, they would be decribed in the same way as graphs are: by precisely naming their 'vertices' (or 'nodes') and the 'edges' connecting them.

Graph minors, on the other hand, the objects that Robertson and Seymour set out to study, are fuzzier substructures than subgraphs: a highly connected minor will usually persist even if the graph containing it is changed a little. Rather than describing these minors in the traditional, somewhat pedestrian, way of naming all their vertices and edges, Robertson and Seymour thought of an ingenious indirect way to capture just their essence: by declaring for every bottleneck in the graph on which of its two sides *most* of that minor lies.[2]

Such a collection of pointers at the bottlenecks of a graph came to be called a *tangle*. Of course, this is a hugely abstract kind of thing – if indeed it merits being called a 'thing' at all. However, bundling even the most complicated collections of objects and their relationships into a single notion is a process not uncommon in mathematics: it enables us to move on and describe more concisely any higher-level structures in which such composite objects occur. In our example, the collection of pointers that constitute a tangle in a graph, one at each of its bottlenecks, deliberately ignores the detail of what vertices and edges our highly connected minor consists of. Instead, it just records where most of it lies – relative to *every* bottleneck.

It turned out that this deliberate restriction of information about the highly connected minors in a graph came with a gain in clarity: the detail discarded was clutter, the information retained its essence.[3]

This development in graph theory was followed by a discovery which, quite unexpectedly, made the entire theory of graph tangles available for the analysis of highly cohesive substructure far beyond graph theory: it turned out that, not just for the notion of a tangle but also for the proofs of the deepest theorems about them, it is enough to know the relative position of those bottlenecks, rather than how exactly they divide the graph of which they are bottlenecks. This information can be encoded in some abstract way that is quite independent of graphs.

> *The theory of tangles has thus become applicable to a wealth of real-world scenarios. The purpose of this book is to show how this can work.*

Our narrative starts with a naive discussion of what tangles mean in various real-world scenarios, and how tangle theory can make an impact there. It then takes the reader through the basic mathematical underpinnings of abstract tangle theory, just enough to enable them to set up a rigorous quantitative framework for applying tangles to their own field. It winds up by revisiting the example scenarios to show how the more formal theory plays out in these contexts.

It should be stressed that those real-world scenarios discussed are highly simplified: they are toy examples of contexts in which tangles can be applied. In reality, they can probably be applied somewhere in most of the natural and quantitative social sciences. This will require the input of experts in those fields. It is the aim of this book to put such experts in a position to try this out for themselves; generic software for this purpose is available via `tangles-book.com`.

The layout of this book is as follows. It begins in Chapter 1 with three short introductions to what tangles are, and what they are designed to achieve: in the natural sciences, in the social sciences, and more specifically in data science. These introductions can be read independently of each other, and are written so as to appeal to readers with these respective backgrounds. In this way, they provide three separate entry points to this book. However, they show aspects of the same big picture, and none of them requires any expertise in the area for which it was written. Hence readers with any background may well benefit from reading all three of them: they are all short, and they illuminate the notion of tangles from three rather different angles.

Chapter 2 develops the notion of a tangle from the intuitive picture formed in Chapter 1, still at an informal level. This will be accomplished by the end of Section 2.3. At this point, any reader who cannot wait to see some tangle applications will be sufficiently equipped to skip ahead to Part II, where applications are discussed informally on the basis of just the notion of tangles, not their theory as described later in the book.

Chapter 3 gives a first indication of the two main theorems about tangles, still not in formal mathematical language but in terms of the example settings described in the three introductions. Together, Chapters 1–3 form Part I of the book, an informal introduction to the notion and theory of tangles from three rather different application perspectives.

Part II continues with a collection of explicit example scenarios in which tangles might be applied, and describes informally what the mere notion of a tangle can already achieve there. As pointed out earlier, these example scenarios are highly simplified and, in their simplicity, artificial. The idea behind going through such a range of examples is to indicate the potential of tangles throughout the sciences, and to do so in an unassuming way that inspires readers to find tangles in their own field of expertise.

The examples in Part II were chosen to illustrate the diversity of potential tangle applications. The corresponding chapter sections may be dipped into at liberty: nothing here is required reading for any material later in the book, except for the corresponding sections in Part IV.

Part III then explains tangles a little more formally. It still does not assume any knowledge of advanced mathematics, but the description is in basic mathematical terms such as sets, subsets, functions and so on. The idea is that this more formal description of the notion of tangles, given in Chapter 7, should be precise enough to enable the reader to apply tangles to their own individual background.

Chapter 8 continues with statements of the two main tangle theorems. The first of these describes how the tangles of a large dataset lie with respect to each other: how some tangles refine others, and how the most refined tangles are separated by some particularly crucial bottlenecks which, between them, organize the dataset into a tree-like shape that displays where its main tangles lie. The second fundamental tangle theorem, which is equally important, tells us what our data looks like if it has no tangles. It offers verifiable quantitative evidence of the *lack* of structure in our data – for example, if it is polluted or inconclusive for some other reason.

The remainder of Part III describes the mathematical toolkit that enables us to tune tangles to fit an intended application (Chapters 9–10), and then describes the fundamental tangle algorithms in Chapter 11.

In Part IV, finally, we return to the examples discussed informally in Part II. Equipped with the formal notions from Part III, and having met the two main tangle theorems, readers will be able to see not just what tangles mean in those various contexts, but also how they can be structured and fine-tuned to offer insights relevant to that field.

Throughout the text, there are markers for 'footnotes' that are collected together at the end of the book. The reason I have implemented these as endnotes is that they can happily be skipped at first reading: they offer further illustrations, more detailed explanations and so on, which are not meant to interrupt the flow of reading unless the reader feels curious for more at that point already.

This book would not exist but for the inspiration and contributions in substance I received from numerous people over the past few years. The development of abstract tangle theory that underlies the applications envisaged here began with an idea of my student Fabian Hundertmark, who extracted from our then recent proof [6] of a canonical tree of tangles theorem for graphs the algebraic core of tangles that was actually needed in that proof [26]. When Sang-il Oum visited me in 2013, we found a proof also of the tangle–tree duality theorem based just on these minimalist algebraic foundations for the notion of a tangle [14, 15]. This set the scene for the development of abstract tangle theory based on [9], which was carried through in the following years mostly by various members of my Hamburg group, particularly by Sandra Albrechtsen, Johannes Carmesin, Christian Elbracht, Ann-Kathrin Elm, Raphael Jacobs, Paul Knappe, Jakob Kneip, Max Teegen, Hanno von Bergen and Daniel Weißauer.

The idea to use this abstract theory of tangles for applications outside mathematics was born when I told Geoff Whittle about it in Oberwolfach in 2016. I remember vividly his exclamation, '*surely*, as we can see structure and things in images so quickly, our brain just sees tangles!'. We then proved that in [16], most of which is now part of Section 14.6.

In the years that followed I benefited immensely from discussing tangle applications with quite a diverse set of people. Outside mathematics these include Partha Dasgupta in economics, Jane Heal in philosophy, Thomas Günther in virology, Chin Li in psychology, the CNRS group around Oliver Poch and Julie Thompson in protein sequencing, Rolf von Lüde in sociology, Ulrike von Luxburg in machine learning, as well as the people at Google including, in particular, Krzysztof Choromanski. Within mathematics they include Nathan Bowler, Joshua Erde, Jim Geelen, Rudi Pendavingh, and Geoff Whittle. To all these I extend my thanks for their ideas, enthusiasm and encouragement!

Last but not least, I thank my tangle software group of Dominik Blankenhagen, Michael Hermann, Fabian Hundertmark and Hanno von Bergen for their amazing success in bringing this pie down from the sky and rooting it firmly in fertile earth. Their generic tangle software is now available via `tangles-book.com` under an open-access licence [1]: for all who would like to play with it or just see some examples in action, to apply it in their own professional context, or to develop it further by adding their own packages to the library.

Reinhard Diestel, February 2024

# Part I

# Tangles

## *A new paradigm for clusters and types*

This first of the four main parts of the book offers a gentle and informal introduction to tangles.

We set out, however, not from what tangles are, but from where they might take us: what difference they might make to some fundamental methodological approaches in the natural sciences, in the social sciences, and in data science. Chapter 1 offers three separate introductions for readers from these three backgrounds. They can be read independently, and in any order. Readers are encouraged to read them all. For not only do they reflect the diversity of potential tangle applications, but through this diversity also highlight seemingly unrelated aspects of the notion of a tangle, the concept central to this book.

The notion of a tangle is then developed informally in Chapter 2. There will be ample reference back to the introductions from Chapter 1, to relate the rather abstract concept of tangles, as it is slowly developed, to their potential applications right away. Readers keen to see some concrete examples of how tangles might be applied, but less curious about the various aspects of the notion as such, may skip ahead to Part II after Section 2.3.

Chapter 3 offers a glimpse of what tangle *theory* has to offer in addition to just the notion of a tangle. The latter, however, goes a long way towards many tangle applications already. Chapter 3 can therefore be skipped by readers who feel sufficiently dedicated to read about tangle theory and its uses in Part III, where both the notion and the theory of tangles are developed rigorously at their simplest mathematical level.

# 1
# The idea behind tangles

This chapter offers three introductions to the concept and purpose of tangles: one for the natural sciences, one for the social sciences, and one for data science. These introductions can be read independently, and readers may choose any one of them as an entry point to this book, according to their own background.

However as all three introductions illuminate the same concept, readers from any background are likely also to benefit from the other two viewpoints. Indeed, while each of them may seem plausible enough on its own, they are rather different. The fact that they nevertheless describe the same concept, that of a tangle, illustrates better than any abstract discussion the breadth of this concept and its potential applications, including in fields not even touched upon here. Moreover, even in a given context where one of the three viewpoints seems more fitting than the other two, switching to one of those deliberately for a moment is likely to add insight that would otherwise be easy to miss.

## 1.1 Tangles in the natural sciences

Suppose we are trying to establish a possible common cause of some set of similar phenomena. To facilitate this, we may design a series of measurements to test various different aspects of each of these phenomena.[1]

If we already have an overview of all the potential causes, we might try to design these measurements so that each potential cause would yield a list of expected readings, one for each of these different measurements, so that different potential causes differ in at least one measurement.

Then only the true cause would be compatible with all the readings we get from our actual measurements performed on the phenomena we are trying to explain.

In our less-than-ideal world, it may not quite work like this. For a start, we might simply not be aware of all the potential causes – not to mention the fundamental issue of what, if anything, is a 'cause'. Similar phenomena may have different causes, or no single cause. Even potentially single causes need not be mutually exclusive but may be able to co-exist; then we shall not be able to design experiments that will exclude all but one of them with certainty. And finally, measurements may be corrupted, but we may not know which ones were.

We usually try to compensate for this by building in some redundancy: perhaps by taking more measurements, or by measuring more different aspects. Or we might resign ourselves to making claims only in probability – which will protect us from being disproved by any single event, but which may also increase immensely the overheads needed to justify precise quantitative assertions (of probabilities).

> *Tangles offer a structural, rather than probabilistic, way to afford the redundancy needed in such cases. They allow us to derive predictions from our data as we would expect them from identifying causes, while sidestepping the philosophical issue of what constitutes a cause.*

The idea, at a high level, is to replace the search for 'causes' with a search for something we can observe directly: structure in our data that occurs *as a result* of the presence of an underlying cause – a kind of structural footprint in observable data that we find whenever phenomena have some common cause, no matter what that cause may be. Different causes have different footprints, but all causes have the same *structural type* of footprint: tangles. The idea is that the structural footprint of each particular cause should carry enough information to replace any reference in the scientific process to that cause, e.g. in making predictions, with a reference to this structure, the tangle that reflects this cause in observable data. If desired, we may think of tangles as an extensional definition of 'cause' that achieves precision and observability at the expense of the intuitive appeal of our informal notion of 'cause'.

In our generic example, a tangle would be a set of *hypothetical* readings for the measurements we have performed on our phenomena, a set of one possible reading per measurement. It would not be just any such set, but one that is typical for the actual readings we got on our phenomena:

> *A tangle is a typical set of measurement readings, one for each measurement, such as a set of readings due to a particular cause.*

It may happen that one, or several, of our phenomena produced exactly this set of readings. But it can also happen that an 'abstract' set of readings is typical, and hence a tangle, for our collection of phenomena without occurring exactly in any one of them. This might be the case, for example, for a set $\tau$ of measurement readings produced by some given cause under laboratory conditions. This set of readings will be typical for our phenomena if they were indeed triggered by this cause, even if none of them produced exactly $\tau$ under our measurements.

What, then, does it mean that $\tau$ is typical for our phenomena? We shall address this question in detail below. Its most important aspect, however, is that our definition of 'typical' will *not* stipulate the existence of some common cause for our phenomena. It will be such that a single cause, or some fixed combination of causes, will produce a set of readings that satisfies our definition of 'typical'. But the definition itself will refer only to our data: the measurement readings we obtained on the phenomena we are seeking to explain. This will allow us to identify tangles directly in the data, without guessing at possible causes, and then *from* these tangles infer that, perhaps, some known cause is present.

Just as a set of similar phenomena can have several possible causes, our measurements might indicate the presence of a single tangle, or several, or none. Given any one of these tangles, we may try to find a common cause for this typical set of readings, or choose not to try. If there *is* a common cause for sufficiently many of the phenomena investigated, it will show up as a tangle and can thus be identified.

But there can also be tangles that cannot, or not yet, be 'explained' by a common cause. Such tangles are just as substantial, and potentially useful, as those that can be labelled by a known common cause; indeed perhaps more so, since the absence of an obvious common cause may have left them unidentified in the past. In this sense, identifying tangles in large sets of similar phenomena can lead to the discovery of new metaphenomena that had previously gone unnoticed and might, henceforth, be interpreted as a 'cause' for the group of phenomena that gave rise to this tangle.

So when is a set $\tau$ of hypothetical measurement readings deemed 'typical' for the actual measurements taken on our phenomena, and is therefore a tangle? There are two notions of 'typical' that are important in tangle theory: a strong one that is satisfied by many tangles but not required in their definition, and a weaker one that is required in their definition, and which suffices to establish the main theorems about tangles.

The strong notion, which we might call *popularity-based*, is that our set of phenomena has a subset $X$ – not too small – such that, for every measurement made, some healthy majority – more than two thirds – of the phenomena in $X$ yielded the reading specified by $\tau$ for that measurement.[2] Note that these may be different two thirds of $X$ for different measurements: every phenomenon, even one in $X$, may for *some* measurements produce readings different from the readings that $\tau$ specifies for those measurements. But every entry in $\tau$ reflects some aspect of what we measured that many of the phenomena in $X$ have in common, and is in this sense typical for all the phenomena in $X$. Clearly, there can be several such tangles $\tau$, witnessed by different sets $X$ of phenomena.

The weaker notion of when our set $\tau$ of hypothetical measurement readings is 'typical' for the actual readings obtained from our measurements, and hence constitutes a tangle, might be called *consistency-based*. It requires of $\tau$ that, for every set of up to three of our measurements, at least one of the phenomena we measured gave the readings specified by $\tau$ for these three phenomena.[3] Note that if $\tau$ is typical in the popularity-based sense it will also be typical in this consistency-based sense,[4] but not conversely.

We often strengthen these consistency requirements of a tangle by asking a little more: that, for some 'agreement parameter' $n$ we chose for the given context, every three measurement readings specified by $\tau$ must be shared not only by one of the phenomena we measured but by at least $n$ of them. Such sets of hypothetical measurement readings, one for every measurement, will thus be even more typical for our phenomena, the more so the larger $n$ is.

All three of these notions of 'typical' are robust against small changes in our data. This makes tangles well suited to 'fuzzy' data with the kind of inherent variation indicated earlier. But the definition of tangles will be completely precise: a formal description of the structure of our data *including* any aspects of its fuzziness. We shall therefore be able to use tangles in rigorous mathematical analysis of our data as it comes.

## 1.2 Tangles in the social sciences

Suppose we run a survey $S$ of political questions on a population $P$ of a thousand people. If there exists a group of, say, a hundred like-minded people among these, there should be a way of answering all the questions in $S$ that is typical for those one hundred people: that, after all, is what 'like-minded' means. Let us write $X$ for this set of a hundred people, and $\tau$ for the way of answering $S$ that is typical for them. Thus, $\tau$ is one typical set of answers to all the questions in $S$.

Let us try to quantify this last assertion, that our answer set $\tau$ is typical for the people in $X$. One way to do this is to require that, for every question $s$ in $S$, some healthy majority – more than two thirds, say – of the people in $X$ agree with the answer to $s$ that $\tau$ specifies. Note that *which* two thirds of $X$ these are may differ for different questions $s$. Every answer specified by $\tau$ reflects the views of a large majority of the people in $X$, and is typical for all of $X$ in this way. But we may not be able to pin down anyone in $P$, let alone two thirds of the people in $X$, that answered all of $S$ as specified by $\tau$.

We shall call such a complete collection $\tau$ of views that is typical for some of the people in $P$ a *mindset tangle*. There may be more than one mindset tangle for $S$, or none, just as there may be several groups of like-minded people in $P$, or none, each with their own typical way of answering $S$.

Traditionally, mindsets are found intuitively: they are first guessed, and only then established by quantitative evidence, perhaps even from a survey designed specifically to test them. Mindset tangles can do that, too. For example, we might feel that there is a 'socialist' way $\sigma$ of answering our survey $S$. We could write these answers down without looking at the actual returns for $S$, just appealing to our intuitive notion of what a 'socialist' way to answer $S$ would be. To test our intuition that this is indeed a typical mindset for the people in $P$ we might then check whether, in the actual returns for $S$ we received from the people in $P$, we can find a sizeable subset $X$ of $P$ as earlier for this particular $\tau = \sigma$.

But tangles can also do the converse: we can identify mindsets, as tangles, in the returns for $S$ without having to guess their answers first:

> *Tangles offer a precise and quantitative way to test for suspected mindsets in a population and to discover unknown ones.*

For example, tangle analysis of political polls in the UK in the years before the Brexit referendum might have detected the existence of an unknown mindset of voters across the familiar political spectrum that helped establish the surprise majority for Brexit in 2016. And similarly in the US with the MAGA[5] mindset before 2016, or that of conservative Greens in Europe in the early 1970s. Tangles can identify previously unknown patterns of coherent views or behaviour.

## 1.3 Tangles in data science

One of the most basic, and at the same time most elusive, tasks in the analysis of big datasets is *clustering*: given a large set of points in some space, one seeks to identify within this set a small number of subsets, called 'clusters', of points that are in some sense similar. If we visualize similarity as distance, clusters will be sets of points that are, somehow, close to each other.

Figure 1.1 shows a simple example of points in the plane. In the picture on the left we can clearly see four clusters. Or can we? If a cluster is a set of points that are pairwise close, and the two points shown in green in the right half of the picture lie in the same cluster, should not the two red points – which are much closer – lie in a common cluster too?

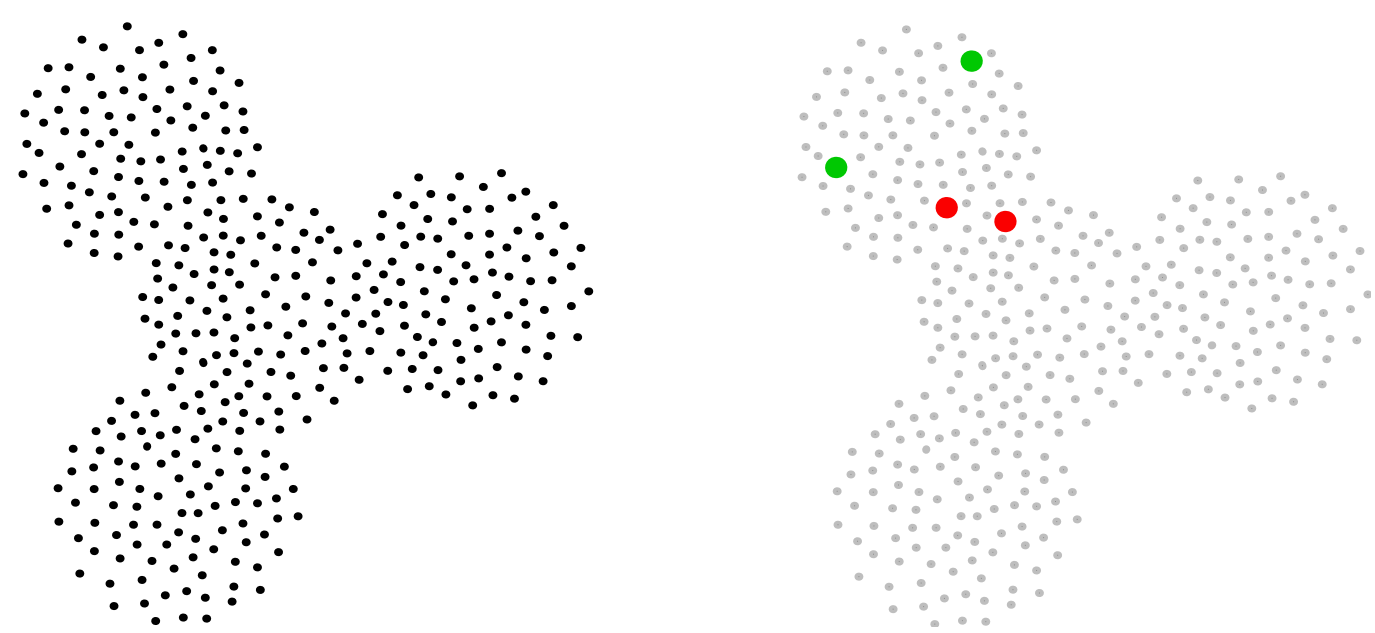

FIGURE 1.1. Four clusters?

For reasons such as this, and other more subtle ones, there is no universal notion of cluster in data science. In our example there are 'clearly' four clusters – but it is hard to come up with an abstract definition of 'cluster' that is satisfied by exactly four sets of points in Figure 1.1, let alone four sets resembling those that we intuitively see as clusters.

Tangles seek to describe clusters in an entirely different manner. Not by dividing the dataset into subsets in some clever new way, but without dividing it up at all: although there will be four tangles in our picture, these will not be defined as sets of points. In particular, questions such as whether the green points should end up in the same cluster but the red points, perhaps, should not, do not even arise.

By avoiding the issue of assigning points to clusters altogether, tangles can be precise without making arbitrary and unwarranted choices:

*Tangles offer a precise, if indirect, way to identify fuzzy clusters.*

Rather than looking for dense clouds of data points, tangles look for the converse: for obvious 'bottlenecks' at which the dataset naturally splits in two. We call ways of splitting our dataset into two disjoint subsets *partitions* of the set, and the two subsets the *sides* of the partition.

Figure 1.2 shows three partitions of our point set at bottlenecks.[6] Now, whatever formal definition of 'cluster' one might choose to work with, one thing will be clear: no bottleneck partition will divide any cluster roughly in half, since that should violate either the definition of a cluster or that of a bottleneck. For example, given one of the three bottlenecks in our picture, and one of the four obvious clusters, we might argue over a few points about whether they should count as belonging to that cluster or not, or on which side of the bottleneck they lie. But for almost all the points in our picture these questions will have a clear answer once we consider a fixed cluster and a fixed bottleneck, no matter how exactly these may be defined.

Put another way, whichever precise definition of cluster (and of bottleneck) someone chose to work with, each of our four intuitive clusters would lie *mostly* on the same side of any partition at a bottleneck. Let us then say that the cluster *orients* this partition towards the side on which most of it lies. Figure 1.2 shows how the central cluster, no matter how it was defined precisely, orients the partitions at the three bottlenecks in this way. Each of the four clusters assigns its own set of arrows to these same three partitions, and the central cluster orients them all inwards.

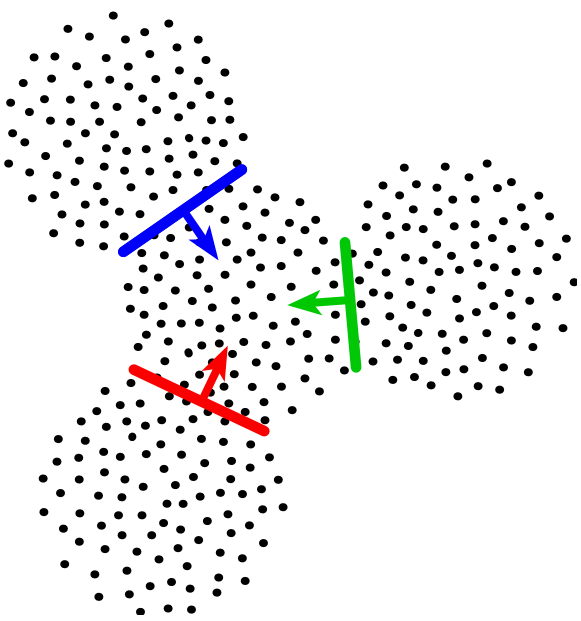

FIGURE 1.2. Orienting the bottlenecks consistently towards the central cluster.

Note that assignments of arrows to bottlenecks that come from one of the four clusters in this way are not arbitrary: the arrows are consistent in that they all point roughly the same way, towards that cluster.

The key idea behind tangles, now, is to keep for each cluster exactly this information – how it orients all the bottleneck partitions – and to forget everything else (such as which points might belong to it). More precisely, tangles will be *defined* as such abstract objects: as *consistent orientations of all the bottleneck partitions* in a data set.

In this way, tangles will extract from the various explicit ways of defining clusters as point sets something like their common essence. Tangles will be robust against small changes in the data, just as they are robust against small changes in any explicit definition of point clusters that we might use to specify them. But their definition as such will be perfectly precise, and involve no arbitrary choices of the kind one invariably has to make when one tries to define clusters as sets of points.

To make this approach work, of course, one has to define formally what the 'bottleneck partitions' of a given dataset are, and when an orientation of all these bottleneck partitions is deemed to be 'consistent'. In our example of Figure 1.2 we defined both these with reference to those four intuitive clusters. Indeed, as 'bottleneck partitions' we took those that split the set of points where this appeared narrow in the picture, which is just another way of saying that we took precisely those partitions that did not cut right through any of the four intuitive clusters; and we called a way of orienting these three partitions 'consistent' if the arrows indicating this pointed towards one of those intuitive clusters.

If we are serious about defining tangles as abstract objects, however, in a bid to bypass the difficulties inherent in trying to define clusters as point sets, then this would beg the question. *The challenge is to define both bottleneck partitions and consistency of their orientations without reference to any perceived cluster*, however vaguely defined. Only once we have achieved this can we 'define' clusters not explicitly as point sets but indirectly as tangles, as is our aim.[7]

To make this challenge a little clearer, let us look at a slightly modified example. Figure 1.3 again shows four clusters with three bottlenecks. This time, one of these has an elongated shape, like a handle. As before, there are clearly four clusters, yet there is no obvious way to define them directly as point sets.

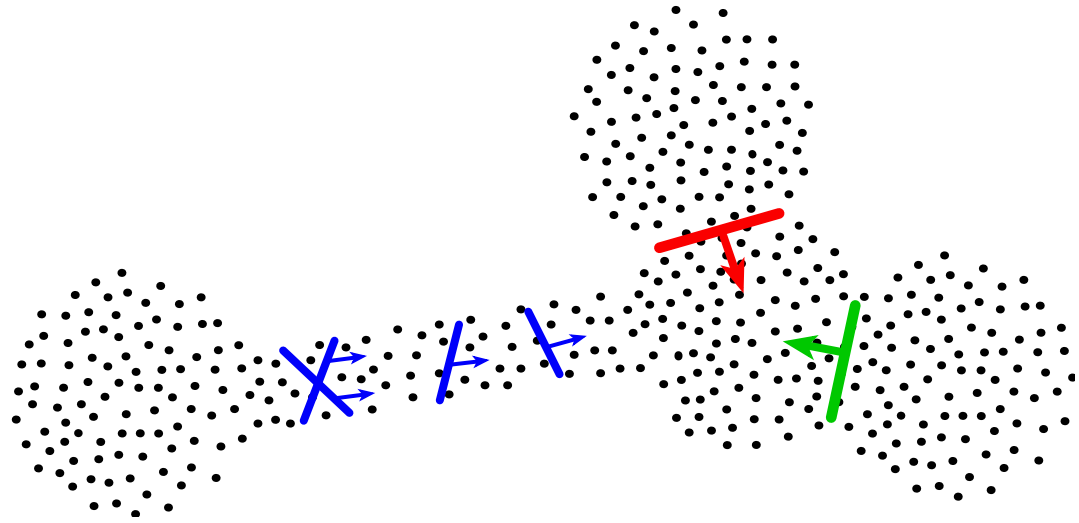

FIGURE 1.3. Three bottlenecks, but many bottleneck partitions, all oriented consistently by the central cluster.

But now there appears to be a problem with our indirect approach too. Intuitively, we would like to orient those bottlenecks; formally we can orient partitions (at bottlenecks), by choosing one of their two sides; but now there are many partitions 'at' the same bottleneck of the handle. Which of these shall we choose to represent that bottleneck?

Since there is no canonical way to make this choice in the abstract, we shall simply consider all those partitions at once: we shall work with 'bottleneck partitions' directly, and find ways to define these formally, but we shall abandon our initial aim to formalize the more intuitive idea of bottlenecks themselves. Our notion of consistency, then, will have to apply also to how we orient partitions at the same bottleneck.

Let us see how this can work in Figure 1.3. We would like there to be four clusters, no more, so only four of the many ways of orienting all its bottleneck partitions should count as 'consistent'. In the picture this can be achieved if, and only if, we can ensure that consistent orientations of partitions at the same bottleneck always point the same way.[8]

In our example, the orientations of bottleneck partitions induced by one of the four obvious clusters satisfy this nicely: any given cluster will either lie mostly on the left of *every* partition at the handle, or mostly on the right of every partition at the handle. Hence, the arrows defined at these partitions by any of our four clusters will either all point to the left, or they will all point to the right, and thus be intuitively consistent.

The challenge remains to come up with a formal definition of consistency as the basis for our notion of tangle that bears this out: one that does not refer to any perceived clusters, but which in the above example will orient all the partitions at the handle in the same direction. Chapter 2 shows how this can be done.

The task of identifying bottleneck partitions in a given dataset with-out reference to intuitively perceived clusters will be our topic later, in Chapter 9. Until then we shall usually assume a set of bottleneck partitions as given.

Once these two tasks are achieved, however, we shall have a definition of tangle which, while being entirely formal and precise, will be able to capture 'fuzzy' clusters in a robust way that does not require us to allocate points to clusters.

## 1.4 On the use of mathematical language in this book

In order to make tangles applicable in such diverse contexts as discussed in this chapter, the language in which we describe them can be based only on what all these contexts have in common. The basic language of modern mathematics is ideally suited to this. It uses only a few fundamental terms: sets and their elements, intersections and unions of sets, perhaps functions that assign to each element of one set some element of another set. And that's all we shall need for the start.

Since this basic mathematical language is so simple, it will be easy to ensure that all definitions and statements made in it will be reliable: they can be taken out of context, referred back to from any point later in the book, and should still stand up to scrutiny. Readers from outside mathematics may find this unusual, and are especially encouraged to rely on it: should they ever find themselves unsure of what some passage later in the book is intended to mean, the solution is likely not just to look out for redundancy near that passage, hoping it is repeated there in slightly different terms, but to look up the definitions of the terms involved.

Another important aspect of mathematical use of language is that, while all notions are precisely defined, the words we use to denote them are usually borrowed from everyday language – like 'tangle'. The reason is that inventing a new word for every new notion would soon make any text unreadable: there are just too many new notions invented all the time. To help with readability, moreover, we try to choose not just any ordinary word for some given new notion, but one that our memory can easily associate with it: a word whose everyday meaning resembles, a little at least, that new mathematical notion. When such a word is then encountered later in the text, it should already be flagged in the reader's mind as having a technical meaning, and it will usually be that meaning, not its everyday meaning, that is meant.

Still, there will remain cases where such ordinary words are still used with their ordinary meaning, and the reader will have to guess from the context when this is the case. A good example is the word 'typical', to which we assigned a technical meaning in Section 1.1 but which we used naively again at the start of Section 1.2. This was necessary, because the later section was meant to be readable as an entry point to this book. But any reader that happened to have read the earlier section first had to guess from the context at the start of Section 1.2 that 'typical' is used there with its everyday meaning.

While we usually hijack everyday words for mathematical notions, individual mathematical objects are often denoted by strange-looking symbols. These are chosen to help the reader remember how these objects relate to others, or how they might be interpreted when tangles are applied. For example, if we chose to denote a bottleneck partition as discussed in Section 1.3 by the letter $s$, it will be natural later to denote its two orientations as $\vec{s}$ and $\overleftarrow{s}$. This is innocuous enough. But remember that $s$, a partition of some dataset $V$, is a pair of subsets of $V$, its two *sides*. Now it may happen in our narrative that one of these sides occurs first, and that we wish to consider $s$ only *because* it has this subset of $V$ as a side. How, then, shall we refer to this side before $s$ is mentioned?

One way to do this would be to introduce that side as a set denoted by some generic symbol for sets, such as $A$, then define $B$ as its complement $V \smallsetminus A$, introduce $s$ as the set $\{A, B\}$, note that this is a partition of $V$ with sides $A$ and $B$, remember that orienting $s$ means picking one of its sides, and finally define $\vec{s}$ as the 'orientation' $A$ of $s$. Apart from the fact that this last step looks notationally odd, introducing the letters $A$ and $B$ is also a lot of notational clutter if all we ever need is $s$ and its two orientations. In such cases we may therefore start our story by giving some interesting subset of $V$ the name $\vec{s}$: just like that, out of the blue.

To non-mathematicians this may look strange, perhaps a little terrifying. But probably only because they suspect some hidden information in this notation which they think they do not know. When this happens, there is a very simple way to keep ones head: to rely on what exactly the text says, and not to worry about the notation. If $\vec{S}$ is introduced as an arbitrary finite set, then that's what it is: a set, nothing special. The reason for calling it $\vec{S}$ rather than $A$, say, may become apparent later,[9] but it will help not to speculate about this too much too early.

The whole point of using mathematical language is that its notions are not overloaded with meaning developed through the centuries, and perhaps subject to debate. Starting our formal description in Part III of what tangles are with just arbitrary sets, their elements and so on, we shall have a clean slate that will be filled only carefully and slowly. Every statement in the text should be understandable, if necessary, on the basis of just the few things said formally before in Part III, so these can be checked if in doubt.

Yet human readers are not machines, and sometimes it helps to have an intuition, however vague, of what some formal notions are intended to describe. Our formal development of tangle terminology will therefore be accompanied by plenty of informal examples.

At the beginning, in Chapter 2, the narrative will in fact be almost entirely in terms of examples. However, in order to become acquainted early with the symbols to be used later in our more formal discourse, we shall introduce these symbols here already: not in a mathematically rigorous way as abstract sets, but naively as sets of ‘things’ that are not mathematical objects. As the book progresses, I shall make more of an effort to keep the levels of formal definitions and of examples visibly separate: so that examples are not formally relied on later, while formal definitions that are meant to be relied on can be easily be identified.

# 2

# The notion of a tangle

The three introductions to tangles given in Chapter 1 focussed not so much on what tangles are but on what they might achieve: to offer a structural and evidence-based alternative to the vague notion of a 'cause' in the natural sciences and to identify these in any given data; a way of recognizing and discovering mindsets in the social sciences; and a new method of clustering in data science.

In this chapter we shift our focus to describing what tangles are. Although we shall frequently refer to the three introductions as background, we shall develop the notion of a tangle again from scratch. In particular, we shall not build on the various notions of 'typical' that we discussed in Section 1.1. Our exposition will still be informal; our mathematically rigorous treatment of tangles will begin in Part III.

Consider a collection $V$ of objects and a set $\vec{S}$ of features[1] that each of the objects in $V$ may have or fail to have. Given such a (possible) feature $\vec{s} \in \vec{S}$, we denote its negation by $\overleftarrow{s}$. The unordered pair $\{\vec{s}, \overleftarrow{s}\}$ of the feature together with its negation is then denoted by $s$, and the set of all these $s$ is denoted by $S$. We may think of these $s = \{\vec{s}, \overleftarrow{s}\}$ as 'potential features': a feature bundled up with its negation into one entity.

For example, if $V$ is a set of pieces of furniture, then $\vec{s}$ might be the feature of being made entirely of wood. Then $\overleftarrow{s}$ would be the feature of being made of any other material, or a combination of materials, and $s$ could be thought of as the question of whether or not a given element of $V$ is made entirely of wood.

In the language of Section 1.1 the elements of $V$ would be the phenomena investigated. The $s \in S$ would be the measurements performed on these phenomena, with two possible outcomes $\vec{s}$ and $\overleftarrow{s}$ (called 'readings' in Section 1.1).

In the example of Section 1.2, the set $V$ would be the population $P$ of people polled by our survey $S$, which for simplicity we assume to consist of yes/no questions. If $\vec{s}$ denotes the 'yes' answer to a question $s \in S$, then $\overleftarrow{s}$ will denote the 'no' answer, and vice versa.

In the clustering scenario of Section 1.3, the set $V$ would be the set of points in which we look for clusters. If we equate a feature $\vec{s}$ with the set of objects in $V$ that have it, then $\vec{s}$ and $\overleftarrow{s}$ form a partition of $V$, the partition $s = \{\vec{s}, \overleftarrow{s}\}$. We may think of $S$ as the set of those partitions of $V$ that are particularly natural, its 'bottleneck' partitions.

## 2.1 Features that often occur together

Tangles are a way to formalize the notion that some features typically occur together. They offer a formal way of identifying such groups of features as 'typical for $V$', each 'type' giving rise to a separate tangle.

Before we make this more precise, let us point out right away that the term 'typical for $V$' will not normally be applied to single features, only to groups of features, which we then call *types*. We think of a group of features as 'typical for $V$' if these features often occur together: if the fact that a given $v \in V$ has one of them makes it more likely that it has the others too – but in a structural, not merely probabilistic, sense.

In order to identify a collection of features as 'typical', however, it will not be necessary to precisely delineate a corresponding set of *objects* (elements of $V$) that have exactly, or even mostly, these features. This reflects most real-world examples, where these sets are at best 'fuzzy'. By working directly on the level of features rather than the level of objects, tangles can be completely precise even when the objects whose features they capture cannot be clearly delineated from each other. This is a particular strength of tangles when they are used for clustering, as indicated in Section 1.3. However, tangles are qualitatively different; they are not just a new clustering method.

Let us return to the example where $V$ is a set of pieces of furniture. Our list $\vec{S}$ of possible features (including their negations) consists of qualities such as colour, material, the number of legs, intended function, and so on – perhaps a hundred or so of possible features. The idea of tangles is that, even though $\vec{S}$ may be quite large, some of its elements may combine into groups[2] that correspond to just a few types of furniture as we know them: chairs, tables, beds and so on.

A key aspect of tangles is that they can identify such types without any prior intuition: if we are told that a container $V$ full of furniture is waiting for us at customs in the harbour, and all we have is a list of items $v$ identified only by an item number together with, for each number, a list of which of our 100 features this item has, our computer – if it knows tangles – may be able to tell us that our delivery contains furniture of just a few types: types that we (but not our computer) might identify as chairs, tables and beds, perhaps with the tables splitting into dining tables and desks.

In the language of Section 1.1 these types would correlate with the different possible 'causes' for objects to be furniture: our need to sit, sleep, eat and so on. In the example of Section 1.2 they would be mind-sets. In the setting of Section 1.3, the sets of chairs, tables and beds would form clusters in $V$. These clusters might not be clearly delineated – for example, if our delivery contains a deckchair – but the types, groups of features that often occur together, would be precisely defined.

In the remainder of this chapter we shall not always make explicit reference to the three example scenarios from Chapter 1. But readers are encouraged to check for themselves what the various new terms mean in each of those contexts, to keep all three aspects alive as they build their intuition for tangles.

## 2.2 Consistency of features

To illustrate how our computer may be able to identify types of furniture from those feature lists without understanding them, let us briefly consider the inverse question: starting from a known type of furniture, such as chairs, how might this type be identifiable from the data if it was not known to be a type?

A possible answer, which will lead straight to the concept of tangles, is as follows. Each individual piece of furniture in our unknown delivery, $v \in V$ say, has some of the features from our list $\vec{S}$ but not others. It thereby *specifies* the elements $s$ of $S$: as $\vec{s}$ if it has the feature $\vec{s}$, and as $\overleftarrow{s}$ otherwise. We say that every $v \in V$ defines a *specification* of $S$, a *choice for each* $s \in S$ *of either* $\vec{s}$ *or* $\overleftarrow{s}$ *but not both.*[3] We shall denote the particular specification of $S$ defined by $v$ as

$$v(S) := \{\, v(s) \mid s \in S \,\},$$

where $v(s) := \vec{s}$ if $v$ specifies $s$ as $\vec{s}$ and $v(s) := \overleftarrow{s}$ if $v$ specifies $s$ as $\overleftarrow{s}$.

Does every specification of $S$ come from some $v \in V$ in this way? Certainly not. There will be no object in our delivery that is both made entirely of wood and also made entirely of steel. Therefore no $v \in V$ will specify both $r$ as $\vec{r}$ rather than $\overleftarrow{r}$, and $s$ as $\vec{s}$ rather than $\overleftarrow{s}$, when $\vec{r}$ and $\vec{s}$ stand for being made of wood or steel, respectively. Now, $S$ has many (abstract) specifications that contain both $\vec{r}$ and $\vec{s}$. But none of these is defined, as $v(S)$, by any real piece $v$ of furniture, because the features $\vec{r}$ and $\vec{s}$ are inconsistent.

Let us turn this manifestation in $V$ of logical inconsistencies within $\vec{S}$ into an extensional definition of 'factual' inconsistency for specifications of $S$ in terms of $V$. Let us call a specification of $S$ *consistent* if it contains no inconsistent triple, where an *inconsistent triple* is a set of up to three[4] features that are not found together in any $v \in V$. Specifications of $S$ that come from some $v \in V$ are clearly consistent, because every three features in $v(S)$ are shared at least by $v$. But $S$ can also have consistent specifications that are not, as a whole, defined by any $v \in V$ as $v(S)$.[5]

Tangles will be specifications of $S$ with certain properties that make them 'typical' for $V$. Consistency will be a minimum requirement for qualifying as typical.

## 2.3 From consistency to tangles

It is one of the fortes of tangles that they allow considerable freedom in the definition of what makes an entire specification of $S$ 'typical' for $V$ – freedom that can be used to tailor tangles precisely to the intended application. We shall discuss this in detail in Chapter 7. But we are already in a position to mention one of the most common ways of defining 'typical', which is just a strengthening of consistency.

To get a prior feel for our (forthcoming) formal definition of 'typical', consider the specification of $S$ in our furniture example that is determined by an 'ideal chair' plucked straight from Plato's heaven: let us specify each $s \in S$ as $\vec{s}$ if this imagined ideal chair has the feature $\vec{s}$, and as $\overleftarrow{s}$ if not.[6] This can be done independently of our delivery $V$, just from our intuitive notion of what chairs are. But if our delivery has a sizeable portion of chairs in it, then this phantom specification of $S$ that describes our ideal chair has something to do with $V$ after all.

Indeed, for every triple $\vec{r}$, $\vec{s}$, $\vec{t}$ of features of our ideal chair there will be a few elements of $V$, at least $n$ say, that share these three features. For example, if $\vec{r}$, $\vec{s}$, $\vec{t}$ stand for having four legs, a flat central surface, and a near-vertical surface, respectively, there will be – among the many chairs in $V$ which we assume to exist – a few that have four legs and a flat seating surface and a nearly vertical back.

By contrast, if we pick twenty rather than three features of our ideal chair there may be no $v \in V$ that has all of those, even though there are plenty of chairs in $V$. But for every choice of three features there will be several – though which these are will depend on which three features of our ideal chair we have in mind.

Simple though it may seem, it turns out that for most sensibly imagined furniture deliveries and reasonable lists $S$ of potential features this formal criterion for 'typical' distinguishes those specifications of $S$ that describe genuine types of furniture from most of its other specifications.[7] But in identifying such specifications as 'types' we made no appeal to our intuition, or to the meaning of their features.[8]

So let us make this property of specifications of $S$ that describe 'ideal' chairs, tables or beds into a more formal, if still ad hoc, definition of 'typical': let us call a specification $\tau$ of $S$ *typical* for $V$ if for every set $R$ of up to three elements of $S$ there are at least $n$ elements $v$ of $V$ that specify $R$ as $\tau$ does, i.e., which satisfy $v(s) = \tau(s)$ for every $s \in R$.[9] The integer $n$ here is a fixed parameter on which our notion of 'typical' depends, and which we are free to choose. We allow $n = 1$, in which case 'typical' means no more than 'consistent'. But as we make $n$ larger, the resulting notion of when a specification of $S$ is 'typical for $V$' gets stronger and stronger.

Crucially, this definition of 'typical' is purely extensional: it makes no reference to what a typical specification of $S$ is 'typical of'. Specifications of ideal chairs, tables or beds are all typical in this same sense: they all satisfy the same one definition.

Equally crucially, a specification of $S$ can be typical for $V$ even if $V$ has no element that has all its features at once. Thus, we have a valid and meaningful formal definition of an 'ideal something' even when such a thing does not exist in the real world, let alone in $V$.

Relative to our notion of 'typical', which is subject to change as we increase $n$, we can now define tangles informally:

*A tangle of $S$ is any specification of $S$ that is typical for $V$.*

Since our ad hoc definition of 'typical' is phrased in terms of small subsets of $\vec{S}$, sets of size at most 3 (of which there are not so many), we can compute tangles without having to guess them first. In particular, we can compute tangles of $S$ even when $V$ is 'known' only in the mechanical sense of data being available (but not necessarily understood), and $S$ is a set of potential features that are known, or assumed, to be relevant but whose relationship to each other is unknown.

Tangles therefore enable us to *find* even previously unknown 'types' of features in the data to be analysed: combinations of features that occur together significantly more often than others. This was important in all three of the scenarios from Chapter 1: tangles can identify previously unknown causes, mindsets, or clusters.

When we treat tangles mathematically in Part III, we define them more broadly as consistent specifications of $S$ that have no subset in some collection $\mathcal{F}$ of subsets of $\vec{S}$, which we may specify as we wish. If we choose $\mathcal{F} = \emptyset$, then all consistent specifications of $S$ will be tangles. Our informal definition of 'typical' then corresponds to choosing $\mathcal{F}$ as

$$\mathcal{F}_n := \{\, \{\vec{r}, \vec{s}, \vec{t}\} \subseteq \vec{S} : |\vec{r} \cap \vec{s} \cap \vec{t}\,| < n \,\},$$

where $|\ |$ denotes the number of elements of a set and the features $\vec{r}$, $\vec{s}$, $\vec{t}$ are interpreted as the subsets of $V$ that have them. So their intersection is precisely the set of all $v \in V$ that have all three features $\vec{r}$, $\vec{s}$ and $\vec{t}$.[10]

A tangle of $S$, then, with 'typical' defined as earlier, or more formally as having no subset in $\mathcal{F} = \mathcal{F}_n$, is any specification of $S$ such that every three of its features are shared by at least $n$ elements of $V$. We call this $n$ the *agreement* value required of these tangles; the variable $n$ in the definition of $\mathcal{F}_n$ is its *agreement parameter.* In general, a tangle of $S$ will be any consistent specification of $S$ no subset of which is an element of $\mathcal{F}$, for any collection $\mathcal{F}$ of subsets of $\vec{S}$ of our choice.

With this notion of a tangle, any reader who cannot wait to see some potential tangle applications is well equipped to skip straight ahead to Part II now, which requires no more knowledge about tangles than their definition. The remainder of Part I offers an introduction to what tangle theory has to offer, which is treated more formally in Part III and will form the basis of our more detailed look at tangle applications in Part IV.

## 2.4 Principal and black hole tangles: two simple examples

In this section we introduce two simple, if somewhat extreme, examples of tangles that will crop up, again and again, throughout the book – often as tangles we want to watch out for because they are not the kind of tangles we are really interested in.

Assume first that, as in Section 1.3, our set $S$ of potential features consists of partitions of $V$ into two non-empty subsets. Every feature $\vec{s}$, then, is one of the two sides of a partition $\{A, B\}$ of $V$: either $A$ or $B$. If $\vec{s} = A$ then $\overleftarrow{s} = B$, and vice versa. In particular, every feature is formally a subset of $V$, and every specification of $S$, including any tangle, will be a set of subsets of $V$.

As our first example, consider for any given $v \in V$ the specification

$$\tau_v = \{\, \vec{s} \in \vec{S} \mid v \in \vec{s}\,\} = v(S)$$

of $S$. These are the *principal* specifications of $S$, one for every $v \in V$. Note that they are consistent, since every $\vec{s} \in \tau_v$ contains $v$.

If $\mathcal{F} = \emptyset$ in our formalization of tangles at the end of Section 2.3, these consistent specifications $\tau_v$ of $S$ will be tangles; we call them the *principal tangles* of $S$. Let us look at a particularly common example.

We say that a specification $\sigma$ of $S$ is *focussed on* $v \in V$ if it contains the singleton subset $\{v\}$ of $V$ as an element. If $\sigma$ is consistent, then this implies that $\sigma = \tau_v$: given any $\vec{s} \in \vec{S}$ such that $v \in \vec{s}$ (as in the definition of $\tau_v$), we cannot have $\overleftarrow{s} \in \sigma$, since $\overleftarrow{s}$ is inconsistent with $\{v\} \in \sigma$, so $\vec{s} \in \sigma$. Hence $\sigma$ specifies every $s \in S$ as $\tau_v$ does, which means that $\sigma = \tau_v$.

Thus, if $S$ has a tangle focussed on some $v \in V$, then this is its principal tangle $\tau_v$. Conversely, $\tau_v$ need not be focussed on $v$, even if it is a tangle, because $\{v\}$ may fail to be an element of $\vec{S}$. But $\tau_v$ cannot be focussed on any $u \neq v$ either, since $\{u\} \notin \tau_v$ by definition of $\tau_v$.

Let us show that these $\tau_v$ are the *only* consistent specifications of $S$, and hence its only possible tangles, when $S$ is not a set of particularly natural 'bottleneck partitions' of $V$ but consists of all the partitions of $V$. Since $\{\{v\}, V \setminus \{v\}\}$ lies in this $S$, the principal specification $\tau_v$ of $S$ will then contain $\{v\}$, and hence be focussed on $v$.

To prove that the $\tau_v$ are the only consistent specifications of the set $S$ of *all* partitions of $V$, consider any such specification $\tau$. Let $\vec{s} = A$ be a smallest element of $\tau$ in terms of $|A|$, the number of elements of $V$ it contains. Note that $A \neq \emptyset$: being consistent, $\tau$ cannot contain the set $\{A, A, A\} = \{A\}$ if the intersection of its elements is empty, which it is if $A = \emptyset$. Let us show that $|A| = 1$.

Suppose $A$ has more than one element. Then $A$ has a partition into two non-empty subsets, $B$ and $C$ say. By the minimal choice of $A$ these do not lie in $\tau$, so their complements $\vec{r} := V \setminus B$ and $\vec{t} := V \setminus C$ do. But now $\{\vec{s}, \vec{r}, \vec{t}\} \subseteq \tau$ while $\vec{s} \cap \vec{r} \cap \vec{t} = \emptyset$, contradicting the consistency of $\tau$. Hence our assumption that $A$ has more than one element is false: it consists of a single element, $v$ say.

Let us prove that $\tau = \tau_v$. Given any partition of $V$, one of its two sides must be in $\tau$. In fact it must be the side containing $v$, since the other side has empty intersection with $\{v\}$, so they cannot both lie in $\tau$. Hence $\tau$ specifies every partition in $S$ as $v$ does: $\tau(s) = v(s)$ for every $s \in S$. This completes our proof that $\tau = \tau_v$.

Depending on our choice of $\mathcal{F}$, this has the following consequences for tangles of this extreme choice of $S$. If $\mathcal{F} = \emptyset$, the tangles of $S$ are precisely its consistent orientations, those of the form $\tau_v$. As soon as we forbid singletons $\vec{s} = \{v\}$ as elements of tangles, however, e.g. by taking $\mathcal{F} = \mathcal{F}_n$ with $n > 1$, or by explicitly putting all sets $\{\{v\}\}$ in $\mathcal{F}$, we have no tangles of $S$ at all: since tangles have to be consistent, they can only be of the form $\tau_v$, but we have just ruled those out.

For this reason, we shall not normally consider as $S$ the set of all partitions of $V$, but mostly sets of partitions that divide $V$ in a particularly natural way. Those will be our 'bottleneck' partitions. Exactly which these should be will be a matter for Chapter 9.

Incidentally, our earlier proof – not the fact – that the set of all partitions of $V$ has only the $\tau_v$ as consistent orientations can teach us something we left unproved in Section 1.3: that any tangle $\tau$ of the bottleneck partitions in Figure 1.3 orients all the partitions at the handle in the same direction. The proof is very similar.[11]

Let us now look at the other extreme: that $S$ contains too few partitions of $V$ – e.g., because the 'bottlenecks' we are allowing are too narrow. For example, let us assume that the only partitions in $S$ are those of the form $\big\{\{v\}, V \setminus \{v\}\big\}$, and let $\tau$ be any consistent specification of $S$.

If $\tau$ contains one of the singleton sets in $\vec{S}$, say $\{v\} \in \tau$, then by consistency it cannot contain another such set $\{u\}$ with $u \neq v$, since $\{u\} \cap \{v\} = \emptyset$. It will therefore contain all the sets $V \setminus \{u\}$ for such $u$. All these contain $v$, so $\tau = \tau_v$. On the other hand, as long as $|V| \geqslant 4$, also $\tau = \big\{ V \setminus \{v\} : v \in V \big\}$ is a consistent specification of $S$, a rather 'unfocussed' tangle of $S$.

If we expand the singleton sets $\{v\}$ to larger subsets of $V$ which, intuitively, form 'clusters' of our data, we can generalize this last example to slightly more general examples of the same type, which we shall revisit several times later in this book: the *black hole tangles*.[12]

To define these more formally, let $V$ have $n \geqslant 4$ disjoint clusters, well separated from each other. Let $S$ consist of only the $n$ partitions of $V$ that each have one of the clusters on one side and the other $n-1$ clusters on the other side. Each of the clusters induces a tangle of $S$: the specification of $S$ which, in the language of Section 1.3, orients every partition $s \in S$ towards that cluster. But there is one more tangle: that which orients each of the $n$ partitions in $S$ away from the cluster it separates from the others. This tangle does not correspond to any cluster in $V$: like a black hole at the centre of a galaxy it sits at its void centre, with all the other tangles arranged symmetrically around it (Figure 2.1).

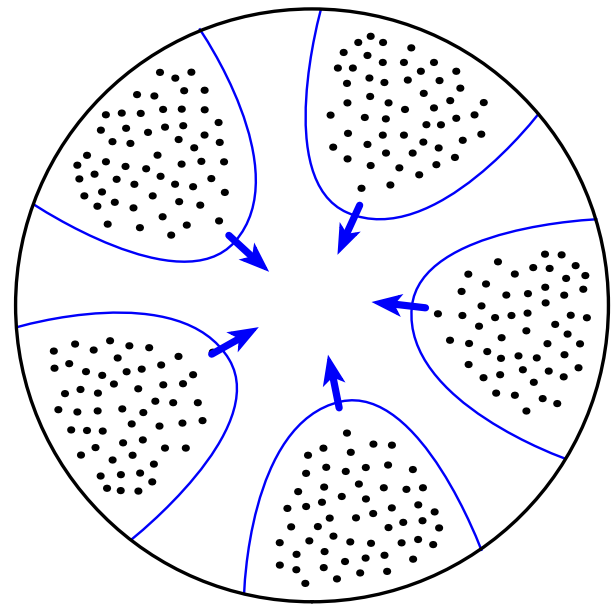

FIGURE 2.1. The *black hole tangle* corresponds to no cluster.

## 2.5 Guiding sets and functions

When we defined a *tangle* of $S$, in Section 2.3, as any specification $\tau$ of $S$ that is typical for $V$, we were assuming a notion of 'typical' that we called *consistency-based* in Section 1.1: for every set $R$ of up to three elements of $S$ there should be at least $n$ elements of $V$ that specify $R$ as $\tau$ does, for some fixed $n \geqslant 1$ of our choice. In the more general definition of tangles at the end of Section 2.3, these tangles were obtained by setting $\mathcal{F} = \mathcal{F}_n$. These are the tangles we consider in this section, with $n = 1$ to allow the most general such tangles.

In Section 1.1 we also discussed another possible notion of 'typical', which we called *popularity-based*. This was that $V$ has a subset $X$, not too small, in which $\tau$ is 'popular' in that for every $s \in S$ more than two thirds of the elements of $X$ specify $s$ as $\tau$ does. We saw that this implies that $\tau$ is typical also in the earlier sense, and hence is a tangle. We may think of $X$ as 'witnessing' this, and as 'guiding' the choices made by $\tau$.

Formally, let us say that a set $X \subseteq V$ *witnesses* that a specification $\tau$ of $S$ is a tangle if every three features in $\tau$ are shared by some $x \in X$. If there are at least $n$ such $x$ for every three features in $\tau$, we say that $X$ *witnesses* that $\tau$ is a tangle *with agreement at least* $n$.

Let us say that $X$ *guides* the specification $\tau$ of $S$ if, for every $s \in S$, there are more $x$ in $X$ that specify $s$ as $\tau$ does than there are $x \in X$ that specify $s$ in the opposite way. If these majorities are greater than 2/3, then $X$ will also witness that $\tau$ is a tangle, no matter how large or small $X$ is. The maximum $p > \frac{1}{2}$ such that, for every $s \in S$, at least $p\,|X|$ of the elements of $X$ specify $s$ as $\tau$ does, is the *reliability* of $X$ as a guiding set, or *guide*, for $\tau$.

In our furniture example, the tangle $\tau$ of being a chair is witnessed and guided by the set $X$ of chairs in $V$ if every feature of our 'ideal chair' $\tau$ is shared by more than two thirds of the set $X$ of all the chairs in $V$. Such guiding sets $X$ were also used in Section 1.2, where we defined a *mindset* as a collection of views established by a political survey $S$ that were 'often held together', in exactly this sense.

More generally, let us say that a 'weight' function $w\colon V \to \mathbb{N}$ *guides* $\tau$ if, for every $s \in S$, the collective weight of all the $v \in V$ that specify $s$ as $\tau$ does exceeds the collective weight of all the $v \in V$ that specify $s$ in the opposite way.[13] If $X \subseteq V$ guides $\tau$, then mapping the elements of $X$ to 1 and all others to 0 is an example of a function that guides $\tau$.

Much of the attraction and usefulness of tangles stems from the fact that, in practice, most of them have such guiding sets or functions [19]. But it is important to remember that the definition of a tangle does not require that such sets or functions exist. It relies only on the notion of consistency, which is defined with reference only to the values of $v(s)$ for the various $v \in V$ and $s \in S$.

In some contexts, however, tangles of $S$ can be defined without any reference to $V$ at all. In our furniture example we could have defined the consistency of a set of features, or predicates, about the elements of $V$ in purely logical or linguistic terms that make no appeal to $V$. Indeed if $\vec{r}$ stands for 'made entirely of wood' and $\vec{s}$ stands for 'made entirely of steel', we said that the set $\{\vec{r}, \vec{s}\}$ is inconsistent. Our definition of this was extensional: that no object in $V$ is made entirely of wood and also made entirely of steel. But we might have appealed instead to the fact that these two predicates are logically inconsistent – which implies that there is no such object in $V$, but which can be established without examining $V$.

The way consistency is defined in general [9], as part of the notion of abstract tangles, is something half-way between these two options: it makes no reference to $V$ but refers only to some axiomatic properties of $\vec{S}$ which reflect our notion that $\vec{S}$ is a set of 'features'. In this way it also avoids any appeal to logic or meaning. Such abstract tangles, however, will not be treated in this book.

For our purposes the only important thing to note about guiding sets or functions is that while many tangles have them, tangles can be identified, distinguished, or ruled out without any reference to such sets or functions. The mindset of being socialist can be identified without having to find any actual socialists, let alone delineating these as a group of people against others. We shall return to guiding sets and functions in Sections 6.1, 14.1, and 14.2. The question of which tangles have guiding sets or functions is studied mathematically in [10] and [19].

# 3

# The two main tangle theorems: an informal preview

Among the many mathematical theorems about tangles there are two fundamental ones that stand out. In terms of potential tangle applications, the first of these shows how tangles structure the dataset of which they are tangles. The second tells us how our data is structured if it has no tangle. Most tangle applications that go beyond finding tangles are based on one of these two results.

Continuing our informal approach from Chapter 2, let $V$ again be a set of objects and $\vec{S}$ a set of possible features of its elements. Tangles are typical specifications of $S$, as defined in Section 2.3.

In order for the two tangle theorems to hold, the set $\vec{S}$ has to be rich enough. If it is not, we can make it so by adding to it some combinations of features already in $\vec{S}$. This will be explained in Chapter 7, Section 7.5. In this chapter we assume that $\vec{S}$ is rich enough in the sense required.

## 3.1 The first tangle theorem: how tangles structure our data

The first main tangle theorem finds a small subset $T$ of $S$ that suffices to distinguish all the tangles of $S$. Recall that every tangle of $S$ is a (typical) *specification* of $S$, a choice of either $\vec{s}$ or $\overleftarrow{s}$ for every $s \in S$. Two different tangles of $S$, therefore, will make different such choices for at least one $s$ – otherwise they would be the same tangle – and we say that every such $s$ *distinguishes* these two tangles. The first of the two main tangle theorems says that we can always find such an $s$ in some small set $T$ which it extracts from $S$.

In the scenario of Section 1.1, this $T$ will be a small set of particular measurements which, between them, suffice to distinguish all the tangles found in the data: for any two types of phenomena, or two causes of similar phenomena, there will be a measurement in $T$ for which these specify different readings.

An expert system built to establish which illness is causing a patient to feel unwell[1] might start by asking for measurements from this set $T$. Since the measurements in $T$ distinguish between all the illnesses, there will be a unique illness compatible with the readings obtained for the measurements in $T$. Moreover, $T$ is, or can be chosen to be, minimal with this property. Once this unique illness has been identified as the likely cause of the patient's ailments, this diagnostic hypothesis can be tested by further measurements particularly relevant to this illness.

In the social sciences scenario of Section 1.2, where $S$ was a survey[2] we have conducted, the set $T$ would be a set of critical questions that suffice to distinguish the mindsets that exist in the population surveyed. For every two mindsets there will be a question in $T$ on which they disagree, and which can thus be used to distinguish them. As an immediate application, $T$ would make a good small questionnaire for a larger study if $S$ was a pilot study designed to test a large number of questions on a smaller population.

In the clustering scenario of Section 1.3 the set $T$ will be a set of partitions, at least one at each bottleneck. For every pair of tangles of the set $S$ of all the bottleneck partitions there will be a partition in $T$ which these two tangles orient differently.

## 3.2 The second tangle theorem: structure when there is no tangle

The second main tangle theorem tells us how our data is structured if it contains no tangle. The non-existence of a tangle in a dataset is a highly relevant and substantial piece of information, which tells us that the data is inconclusive in some objective and verifiable quantitative sense. In the mindsets scenario of Section 1.2, for example, the theorem furnishes inconclusive poll returns with a verifiable proof that mindsets not only were not found but do not in fact exist.

Given a set $S$ that admits no tangle, the theorem returns a small subset $T$ of $S$ that already has no tangle: then $S$ cannot have one either, since it would include one of $T$. Moreover, the theorem produces a small set of ‘witnesses’ to the non-existence of a tangle of $T$, and hence of $S$: a set of no more than $|T| + 1$ triples in $\vec{T}$ that are either inconsistent or in $\mathcal{F}$ (both of which a computer can easily check), and hence cannot be subsets of any tangle of $T$. These triples witness the non-existence of a tangle of $T$ because they cover $\vec{T}$ in such a clever way – which a computer can check even without looking at $V$ – that any tangle of $T$ *would* have to contain one of these triples.

As with the first main tangle theorem, the set $T$ and those triples cannot simply be found by trial and error. In both theorems, they are highly valuable in both senses of the word: knowing them gives us a lot of useful information that we would not otherwise have, and proving their existence requires some nontrivial mathematics. However, now that this has been done, they can often be found quickly by a computer, using the algorithms we shall meet in Chapter 11.

## 3.3 The predictive power of tangles

The general idea of trying to predict a person’s likely behaviour in a future situation from observations of their actual behaviour in some past situations is as old as humanity: as we learn ‘how someone ticks’, we are better able to make such predictions.[3]

If we have the chance to choose, or even design, those earlier situations, e.g. by selecting them carefully from a collection of past situations in which our individual’s behaviour was observed, or by devising a test consisting of hypothetical situations the response to which our individual is willing to share with us, we have a chance of improving our predictions by choosing such past or hypothetical scenarios that are particularly relevant. Tangles can help to identify these.

Let $P$ be a set of people in whose likely actions we are interested, and $\vec{S}$ the set of possible such actions. To keep things simple, let us consider the example from the introduction in Section 1.2: there, $\vec{S}$ is a set of possible views of the people in $P$,[4] and we wish to predict how a person from $P$ would answer the questions from $S$. Let us assume that, as the basis for our prediction, we are allowed to quiz that person on some small set $T$ of questions, and base our prediction for their answers to $S$ on their known answers to the questions in $T$. Our aim is to choose $T$ so as to make these predictions as good as possible.

The first of our two main tangle theorems, discussed in Section 3.1, is designed to produce a particularly suitable set $T$ of questions, one whose answers entail more predictive power for $S$ than an arbitrary subset of $S$ of that size would. This is because $T$ consists of just enough questions to distinguish all the tangles of $S$: the typical ways to answer $S$, the mindsets (regarding $S$) that exist in $P$.

The answers of an individual to the questions in $T$ thus determine exactly one such type, or mindset: there is only one typical way of answering all the questions in $S$ that includes these particular answers to $T$. This typical way of answering $S$, a tangle of $S$, is a an especially good prediction for the answers of our individual to the other questions in $S$ if we assume that he or she follows any mindset at all, which seems more likely than not.

On the face of it, there appears to be a problem with this approach in that we are trying to base our predictions for somebody's answers to the questions in $S$ on being able to compute the tangles of $S$ first, which in turn requires that we already know what we are trying to predict: the answers of the people in $P$ to the questions in $S$. However, this is not in fact a problem. In any real-world application we would compute these tangles based on how the people in some representative subset $V$ of $P$ answered $S$, and then test the individual we are interested in, most likely someone in $P \setminus V$, on the set $T$ computed for these tangles.

In a typical application in the social sciences, $S$ might be a pilot study run on a small subset $V$ of a larger population $P$. Then $T \subseteq S$, once computed from the return of the pilot study, might be used as the main study run on $P$, with fewer but particularly relevant questions selected from $S$. The answers to $T$ of an individual in $P$ can justify predictions on how this individual would answer the rest of $S$, and $T$ is a particularly well chosen subset of $S$ for this purpose.